\documentclass{article}
\usepackage{spconf,amsmath,graphicx}
\usepackage{subcaption}
\usepackage{bm}

\title{Improving Unimodal Inference with Multimodal Transformers}
\name{Kateryna Chumachenko$^{\dagger}$, Alexandros Iosifidis$^{\star}$, Moncef Gabbouj$^{\dagger}$ \thanks{This project has received funding from the European
Union’s Horizon 2020 research and innovation programme
under grant agreement No 871449 (OpenDR).}}
\address{$^{\dagger}$Tampere University, Faculty of Information Technology and Communication Sciences, Finland  \\
    $^{\star}$Aarhus University, Department of Electrical and Computer Engineering,  Denmark 
    }
\begin{document}
\ninept
\maketitle
\begin{abstract}
This paper proposes an approach for improving performance of unimodal models with multimodal training. Our approach involves a multi-branch architecture that incorporates unimodal models with a multimodal transformer-based branch. By co-training these branches, the stronger multimodal branch can transfer its knowledge to the weaker unimodal branches through a multi-task objective, thereby improving the performance of the resulting unimodal models. We evaluate our approach on tasks of dynamic hand gesture recognition based on RGB and Depth, audiovisual emotion recognition based on speech and facial video, and audio-video-text based sentiment analysis. Our approach outperforms the conventionally trained unimodal counterparts. Interestingly, we also observe that optimization of the unimodal branches improves the multimodal branch, compared to a similar multimodal model trained from scratch.
\end{abstract}

\section{Introduction}
\label{sec:intro}

The availability of an abundance of data in the modern world has driven the development of machine learning methods exploiting such data to their fullest. Recently, there has been an increase in emergence of novel approaches utilizing multiple data modalities simultaneously, such as video, audio, text, or other sensor data, for solving a variety of tasks \cite{alayrac2020self, hu2021unit}. Such methods are referred to as multimodal methods and they have been proven successful in a plethora of application fields, including emotion recognition \cite{chumachenko2022self}, hand gesture recognition \cite{zhang2018egogesture}, human activity recognition \cite{mmtm}, and others. Leveraging multiple data sources concurrently can lead to improved performance of the learning model as data of different modalities can complement and enrich each other.

\begin{figure}
\begin{center}
\includegraphics[width=0.475\textwidth]{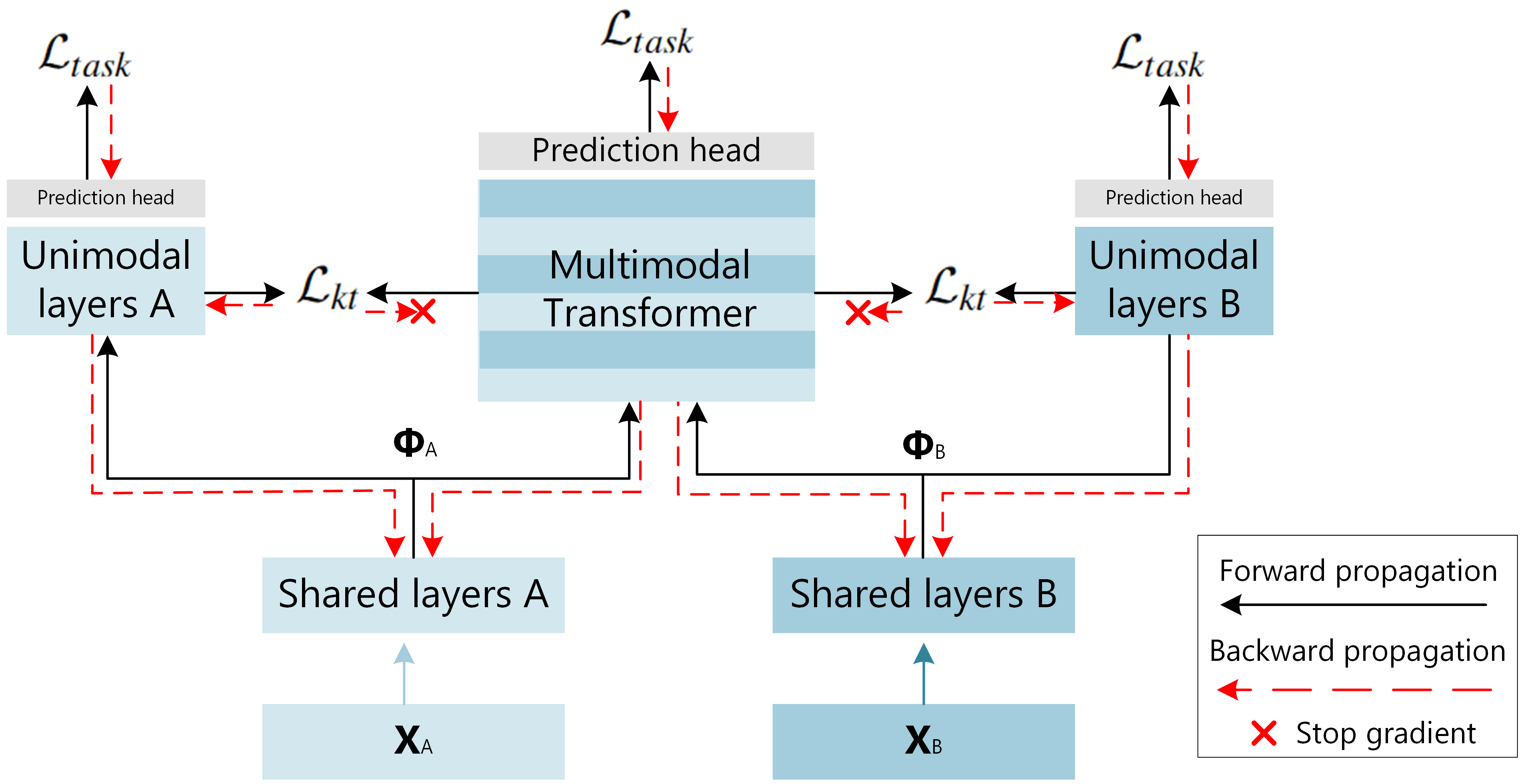}
\end{center}
   \caption{\footnotesize{Description of the proposed framework. For a two modality case A and B, the architecture is comprised of two unimodal branches and a joint multimodal Transformer branch. Early feature extraction layers are shared between multimodal Transformer and corresponding unimodal branches, and both uni- and multimodal branches have their own task-specific heads. Additionally, unimodal branches optimize knowledge transfer criteria from multimodal Transformer, while multimodal branch is not updated based on this criterion. At inference time, multimodal branch is dropped and each of the unimodal branches can be used as a standard unimodal model (alternatively, multimodal branch can be used on its own, too).}}
\label{fig:our}
\vspace{-10pt}
\end{figure}

Research within the field of multimodal methods has been largely focused on tasks where all modalities of interest are assumed to be present both during training and test stages, and has involved development of novel feature fusion methods \cite{mmtm}, solving multimodal alignment problems \cite{mult}, etc. Nevertheless, it is not always desirable to rely on the assumption of all modalities of interest being present at inference time. In real-world applications, data of one or multiple modalities might be unavailable at arbitrary inference steps due to, e.g., transmission delays and media failures, or simply the application at hand might not be suitable for utilizing certain modalities, while they might be available during training. Utilization of unimodal models therefore remains widely adopted due to their simplicity and easier applicability to real-world tasks. Nevertheless, models relying only on unimodal data at inference time can benefit from multimodal training. Such approach can aid in learning richer feature representations from single modality by relating it with other modalities, and help highlight unimodal information that is most relevant for the task. At the same time, the computational costs associated with the model are not increased.

In this work, we propose an approach for improving performance of unimodal models with multimodal training, and employ a multi-branch architecture with both unimodal, and multimodal Transformer-based branches. Unimodal and multimodal branches are co-trained and knowledge from the stronger multimodal branch is continuously transferred to the unimodal branches via a multi-task objective, hence improving the performance of resulting unimodal models. We perform experiments on three multimodal tasks and observe consistent improvements in the performance of the models. At the same time, we also observe that our approach not only improves the performance of unimodal models, but also that of the multimodal teacher model, compared to the similar model trained from scratch. Our contributions can be summarized as follows:
\begin{itemize}
\setlength\itemsep{0em}
\item We propose an approach for improving the performance of arbitrary unimodal models with multimodal training, with no additional computational cost incurred by unimodal model at inference time;
\item The proposed framework is agnostic of the underlying modalities or unimodal architecture types, while in the experiments we showcase various architectures, including 3D-CNNs, 2D+1D-CNNs, and transformer-based ones; 
\item We validate our approach on three multimodal tasks and observe consistent improvements, with different modalities, architectures, and loss functions.
\end{itemize}

\section{Related Work}
Modern research directions in the field of multimodal learning have largely focused on advanced modality fusion methods \cite{hu2021unit, alayrac2020self, baltruvsaitis2018multimodal} and include a variety of approaches, ranging from CNN-based cross-modal Squeeze-and-Excitation blocks \cite{mmtm}, to translation based approaches \cite{lee2021face}.
Within the field of multimodal fusion, perhaps the most notable recent development is the adoption of multimodal Transformers that allow to capture global correspondences between modalities, hence making them an especially favorable choice for temporal sequence modelling tasks where alignment between modalities is an important challenge \cite{chumachenko2022self, icasspav, interspeech}. The idea behind cross-modal Transformers lies in adoption of self-attention mechanism \cite{vaswani2017} with queries and key-value pairs originating from different modalities, and one of the most notable instantiations of such approach is the Multimodal Transformer (MULT) \cite{mult}.

Nevertheless, the above-mentioned approaches have their limitations. Primarily, they all rely on the assumption that the same set of sensors/modalities are available at both training and inference, while such expectation is idealistic and is an especially relevant limitation for real-world applications where flexibility is required. 
A set of methods aim to solve this issue by introducing the multimodal training unimodal testing paradigm, aiming at improving unimodal models by utilizing multimodal data during training. Such methods can be broadly categorized into a few types, with the first type being the methods aiming to reconstruct or otherwise hallucinate a missing modality \cite{garcia2019learning, garcia2018modality, teng2021unimodal, giannone2019learning}. Other methods optimize certain alignment objectives between multiple modalities, e.g., by contrastive learning \cite{meyer2020improving}, or by spatiotemporal semantic alignment \cite{abavisani2019improving}. Nevertheless, such methods are mostly suited for well-paired modality types, such as RGB and Depth, or RGB and Point Clouds, while having limited suitability for modalities where data types are drastically different and their correspondence is not immediately obvious, e.g., audio and RGB frames, or text and RGB frames. In our work, we take aim to overcome this issue, and propose a generalized framework suitable for various data modalities and unimodal architectures.

\section{Proposed approach}
This section describes the proposed approach for improving the performance of an arbitrary unimodal model with multimodal training. We consider the following problem setup: given a set of data representations of arbitrary modalities and corresponding unimodal model architectures, we seek to improve performance of said unimodal models by exploiting multimodal information during training. Concretely, our approach relies on a general framework in which unimodal models are united in a joint architecture by a multimodal Transformer-based branch attached to intermediate features of unimodal models of each modality, hence each unimodal model becomes a separate branch. The multimodal branch is jointly co-trained with resulting unimodal branches, and shares early feature extraction layers with the unimodal branches. Additionally, knowledge transfer between the multimodal Transformer and the unimodal branches is achieved by optimizing a multi-task objective. During inference, the multimodal branch as well as branches corresponding to modalities that are not of interest are dropped, restoring the original architecture of the unimodal model, but with parameters optimized during multimodal training. Overall, a schematic representation of the proposed approach, with two example modalities $A$ and $B$, is outlined in Figure~\ref{fig:our}. 

As can be seen, data of each modality $i$, $\mathbf{X}_i$, is input to a sequence of layers serving as backbone for both unimodal and multimodal branches, resulting in feature representation $\bm{\Phi}_i$ for modality $i$. Further, $\bm{\Phi}_i$ is processed with the remaining part of the unimodal branch, as well as the multimodal Transformer branch (as described further) independently, where each branch has its own task-specific head that optimizes the task-specific objective $\mathcal{L}_{task}$ (e.g., cross-entropy for classification tasks). Additionally, a knowledge transfer objective from stronger multimodal branch to weaker unimodal branches $\mathcal{L}_{kt}$ is optimized, where $\mathcal{L}_{kt}$ can be represented by a variety of different objective functions, as will be discussed further.

Unimodal and multimodal branches as well as task-specific and knowledge transfer objectives are optimized jointly. Shared feature layers receive gradient updates from task-specific objectives of both uni- and multimodal branches, hence forcing them to remain informative for both inference paths and avoiding the loss of modality-specific information, while retaining information useful for modality fusion. 
In turn, knowledge transfer objective encourages the remaining segment of unimodal branch to learn in accordance with the multimodal transformer, hence improving its performance.

\vspace{-8pt}
\subsection{Multimodal Transformer}
\begin{figure}
\centering
\includegraphics[width=0.5\textwidth]{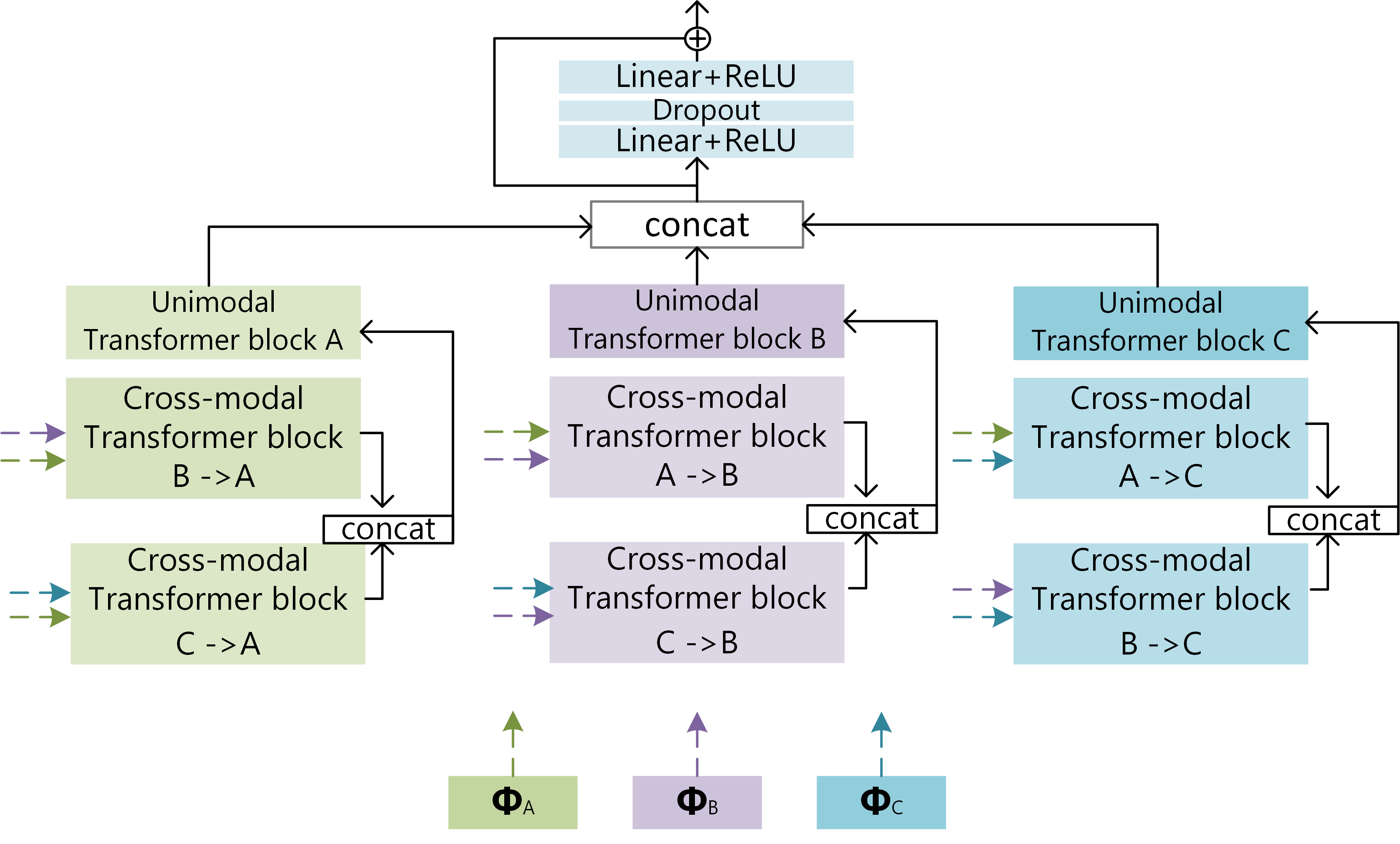}
\caption{\footnotesize Example of a multimodal Transformer with three modalities A, B, and C.}         \label{fig:trans}
\vspace{-18pt}
\end{figure}
Here, we describe the multimodal Transformer branch. Given feature representations of two modalities $\bm{\Phi}_A$ and $\bm{\Phi}_B$, cross-modal attention that fuses modality $B$ into modality $A$ is defined as:
\begin{equation}
\small{
    \hat{\bm{\Phi}}_{AB} = softmax\left(\frac{\mathbf{W}_q\bm{\Phi}_A\bm{\Phi}^T_B\mathbf{W}^T_k}{\sqrt{d}}\right)\mathbf{W}_v\bm{\Phi}_B,}
\end{equation}
followed by another linear projection layer, where $\mathbf{W}_q$, $\mathbf{W}_v$, and $\mathbf{W}_k$ are learnable projection matrices, $d$ is the feature dimensionality, and $\bm{\Phi}_A$ and $\bm{\Phi}_B$ are features of modalities $A$ and $B$. This is generally referred to as cross-attention and it is a generalization of the self-attention mechanism \cite{vaswani2017} where queries originate from modality $A$ and key-value pairs originate from modality $B$. Similarly, fusion of modality $B$ into modality $A$ is achieved by learning queries from modality $B$ and key-value pairs from modality $A$.

The overall multimodal Transformer branch is similar to the one proposed in \cite{mult} and consists of the previously defined cross-attention blocks, optionally followed by unimodal self-attention blocks in each modality, as shown in Figure~\ref{fig:trans}. That is, for fusion of two modalities $A$ and $B$, two cross-attention blocks $A->B$ and $B->A$ are employed and their resulting features concatenated, and in the case where the number of modalities is greater than two, pair-wise cross-attention blocks are calculated within each pair. The prediction head is unimodal model-specific. 

\vspace{-8pt}
\subsection{Unimodal branches}
The proposed approach is agnostic of underlying unimodal models and can be combined with an arbitrary architecture. For the sake of completeness, we describe several examples of architectures used in our experimental evaluation further. For the task of dynamic gesture recognition based on RGB and Depth modalities, each unimodal branch is either an I3D \cite{i3d} or MobileNetv2 \cite{mobilenetv2} architecture, primarily based on 3D convolutional layers. The multimodal branch in I3D variant is attached after $``Mixed\_4f"$ layer, and in the case of MobileNetv2, prior to the last two convolutional blocks. Hence, the majority of the layers is shared between the multimodal and unimodal branches. The extracted 3D convolutional features $\bm{\Phi}$ have the shape of $B\times C\times T\times H \times W$, on which we perform spatial mean pooling, resulting in $B\times C\times T$ input tokens input to the multimodal Transformer.  For the task of audiovisual emotion recognition, we adopt an architecture similar to \cite{chumachenko2022self}, with vision branch being the EfficientNet backbone followed by blocks of 1D-Convolutional layers, and audio branch is also a set of 1D-Convolutional layers. Here, we add multimodal Transformer branch on the output of ``Stage 1'' convolutional block in both branches. This can be compared to `intermediate transformer' fusion described in \cite{chumachenko2022self}, where outputs of multimodal Transformers are not fused back to their corresponding branches, but instead connect to their own output layer.

\vspace{-8pt}
\subsection{Multi-task training objective}
The overall training objective is given by 
\begin{equation}
    \mathcal{L} = \alpha\sum_{i=1}^M{\mathcal{L}_{kt}^i} + \beta\sum_{i=1}^M{\mathcal{L}_{task}^i} + \gamma\mathcal{L}_{task}^{mm},
\end{equation}
where $i$ is the modality indicator, $\mathcal{L}_{task}^i$ is task-specific objective for branch of modality $i$, $\mathcal{L}_{task}^{mm}$ is the task-specific objective of the multimodal branch, and $\mathcal{L}_{kt}^i$ is the knowledge transfer loss from multimodal branch to unimodal branch $i$, and $\alpha, \beta, \gamma$ are scaling coefficients. A multitude of objective functions can serve the purpose of knowledge transfer. Here, we consider three cases, which we refer to as \textit{decision-level alignment}, \textit{feature-level alignment}, and \textit{attention-level alignment}.

In \textit{decision-level alignment} objective, the goal is to transfer high-level information about predictions and class probability distributions from stronger multimodal branch to weaker unimodal branch. To achieve this, for standard classification tasks, we formulate knowledge transfer as knowledge distillation task \cite{hinton2015distilling} and optimize KL-divergence $\mathcal{L}_{kt}^{KL}$ between soft pseudo-labels generated by multimodal branch and softmax outputs of unimodal branches. Soft probability distribution between classes is achieved by applying temperature $T>1$ to predicted class probabilities. Such knowledge transfer allows the unimodal model to capture fine-grained class boundaries from the stronger multimodal model.

In \textit{feature-level alignment} objective, the goal is to transfer broader semantic feature-level information from multi- to unimodal branch. Such formulation can be more general and suitable for a wider variety of tasks. For this goal, we adopt cosine similarity $\mathcal{L}_{kt}^{cos} = \frac{\phi_A \cdot \phi_B}{||\phi_A|| \cdot ||\phi_B||}$ between the final hidden layer output features of the multimodal and unimodal branches, hence promoting the transfer of feature-level semantic information, aimed at improving the performance of task at hand.

Lastly, when unimodal branch architectures are also Transformer-based, a mechanism that we refer to as \textit{attention-level alignment} can be employed. Here, knowledge transfer can be achieved by aligning self-attention probability distributions over temporal tokens in unimodal and multimodal branches. Intuitively, tokens in multimodal Transformer attend to tokens of other modalities globally via self-attention in cross-modal Transformer blocks. Subsequently, unimodal Transformer blocks in multimodal Transformer operate over tokens that have already `seen' corresponding tokens of other modalities. The softmax probabilties of unimodal self-attention in final stages of multimodal Transformer can then be distilled to the corresponding unimodal branches similarly to the first case, by calculating KL-divergence over soft pseudo-labels. We further refer to this approach and objective function as $\mathcal{L}_{kt}^{att}$.

\section{Experimental evaluation}
As described earlier, to the best of our knowledge the few existing methods aimed at unimodal inference with multimodal training are primarily suitable for well-paired modalities as they rely on fine-grained spatial information transfer or modality reconstruction/hallucination. This makes their application in more general scenarios and more heterogeneous modalities largely non-trivial if not impossible. On the other hand, our proposed approach is generalized and makes no assumption on the underlying data. Therefore, to show the effectiveness of our method, we compare the models trained within our framework to unimodal counterparts proposed in recent literature \cite{chumachenko2022self, mult, i3d, mobilenetv2} on a variety of tasks and modalities of different types, and show that our proposed approach improves their performance. We perform experiments on three tasks / datasets: egocentric dynamic gesture recognition using EgoGesture dataset \cite{zhang2018egogesture}, audiovisual emotion recognition using RAVDESS dataset \cite{ravdess}, and multimodal sentiment analysis on CMU-MOSEI dataset \cite{mosei}. We train independently unimodal models with available modalities; multimodal model comprised of shared layers and multimodal Transformer; and the proposed multimodal architecture with knowledge transfer trained jointly, where we evaluate each of the resulting unimodal and multimodal branches independently. In each dataset, we report the performance on the test set, with the model selected based on best performance on the validation set. Each modality model is selected independently from other modalities and knowledge transfer loss weight is a hyperparameter. Best result is highlighted in bold, and results outperforming the baseline are underlined.

\begin{table}[h]
\footnotesize{
\centering
\begin{tabular}{|l|ccc|}
\hline
\textbf{Method}                  & \textbf{Acc-RGB}       & \textbf{Acc-Depth}     &\textbf{Acc-MM}     \\ \hline\hline
MobileNetv2 \cite{mobilenetv2}   & 86.07          & 86.67     & 87.64     \\ 
MobileNetv2-${\mathcal{L}_{kt}^{KL}}$ (ours)   & \textbf{88.57} & \textbf{88.34} & \textbf{89.19}          \\ \hline\hline
I3d \cite{i3d}         & 90.69          &    90.64            &   91.78            \\ 
I3d-${\mathcal{L}_{kt}^{KL}}$ (ours)          & \textbf{91.96} & \textbf{91.84} & \textbf{92.78} \\\hline \hline
\multicolumn{4}{|c|}{\textbf{Ablation studies} } \\ \hline
I3d-${\mathcal{L}_{kt}^{KL}}$, no know. trans.& 90.54 & 90.32 & 92.32 \\
I3d-${\mathcal{L}_{kt}^{KL}}$ (ours) - frozen& 91.74 & 91.82 & 92.73  \\ \hline
\end{tabular}}
\caption{\small Results on EgoGesture dataset.}\label{tab:egogesture}
\end{table}
\textbf{Hand gesture recognition.}
For egocentric dynamic hand gesture recognition, we use EgoGesture dataset \cite{zhang2018egogesture, cao2017egocentric}, which is a hand gesture recognition dataset comprised of RGB and Depth modalities and including 83 hand gesture classes depicted in 24,161 short hand gesture clips, performed by 50 subjects. Unimodal branches are as described in Sec. 3.2, and multimodal branch is comprised of a multimodal Transformer attached to intermediate layers of Depth and RGB branches. As this task is formulated as a video classification problem, we adopt decision-level alignment for knowledge transfer, and minimize KL-divergence with $T=5$ between soft output probability distributions of multimodal and unimodal branches. 

Table~\ref{tab:egogesture} shows the results of the proposed approach. As can be seen, the proposed training framework outperforms the unimodal counterparts on both modalities and both architectures, leading to up to $2.5\%$ improvement in accuracy. Interestingly, we observe that the proposed approach also improves the performance of the multimodal branch when it is trained in conjunction with unimodal branches, compared to the multimodal branch trained independently. This shows that providing unimodal feedback during training forces the shared feature layers to retain more information specific to each independent modality, hence improving the multimodal performance. 

\begin{table}[h!]
\footnotesize
\centering
\begin{tabular}{|l|ccc|}
\hline
     \textbf{Method}        & \textbf{Acc-Audio}      &\textbf{ Acc-Video}     & \textbf{Acc-MM}     \\
             \hline\hline
Unimodal models \cite{chumachenko2022self} & 60.92          & 60.00            & 64.92    \\
MM-${\mathcal{L}_{kt}^{KL}}$ (ours)         & \textbf{63.16} & \textbf{63.16} & \textbf{66.33} \\
\hline
\end{tabular}\caption{\small Results on RAVDESS dataset.}
\label{tab:ravdess}
\vspace{-9pt}
\end{table}
\textbf{Audiovisual emotion recognition.}
For audiovisual emotion recognition we employ the RAVDESS dataset \cite{ravdess} which consists of face and speech recordings of 24 actors acting out 8 emotions and posing a classification task, with 60 video sequences recorded for each actor. The architecture follows the description in Section 3.2, with unimodal models trained from scratch. Knowledge transfer loss $\mathcal{L}_{kt}$ is the KL-divergence between soft outputs with $T=5$ and the task-specific loss is standard cross-entropy. Table~\ref{tab:ravdess} shows the results obtained in audiovisual emotion recognition tasks. As can be seen, the findings are consistent with those obtained in previous task, and the proposed approach improves both unimodal counterparts by up to $3\%$. Similarly, the multimodal branch is improved as well.

\begin{table}[h]
\footnotesize
\centering
\begin{tabular}{|l|ccc|}
\hline
\textbf{Method}              & \textbf{MAE}                 & \textbf{Corr}                & \textbf{Acc\_7}       \\
              \hline\hline
Audio \cite{mult}   & 0.8146          & 0.2395          & 41.05    \\
A-${\mathcal{L}_{kt}^{cos}}$ (ours)    & \underline{0.8125} &\textbf{0.2812}& 40.76  \\ 
A-${\mathcal{L}_{kt}^{att}}$ (ours) & \textbf{0.8111} & \underline{0.2493} & \textbf{41.17} \\ \hline

Vision \cite{mult}   & 0.8079          & 0.2313          & 42.18                \\

V-${\mathcal{L}_{kt}^{cos}}$ (ours)     & \underline{0.8028} & \textbf{0.2774} & \underline{42.18}  \\ 
V-${\mathcal{L}_{kt}^{att}}$ (ours)  & \textbf{0.7978} & \underline{0.2680} & \textbf{42.73} \\ \hline

Text \cite{mult} & 0.6290          & 0.6481           & 48.72            \\

T-${\mathcal{L}_{kt}^{cos}}$ (ours)  & \textbf{0.6199} & \textbf{0.6570} & \textbf{49.62}  \\ 
T-${\mathcal{L}_{kt}^{att}}$ (ours) & \underline{0.6203} & \underline{0.6537} & \underline{49.02} \\ \hline

Multimodal \cite{mult} & 0.6407 &  0.6748 &  48.72  \\
MM-${\mathcal{L}_{kt}^{cos}}$ (ours)  & \textbf{0.6273} & \textbf{0.6793} & \textbf{49.32}  \\
MM-${\mathcal{L}_{kt}^{att}}$ (ours) & \underline{0.6331} & 0.6625 & \underline{49.09} \\
\hline
\end{tabular}
\caption[]{Results on MOSEI dataset.}\label{tab:mosei}
\vspace{-8pt}
\end{table}
\textbf{Multimodal sentiment analysis}
Next, for the task of multimodal sentiment analysis, we perform experiments on the unaligned version of CMU-MOSEI dataset \cite{mosei}, which contains 23,454 utterances extracted from movie review video clips taken from YouTube. The dataset consists of audio, vision, and text modalities, where each utterance is labeled with a sentiment score in the range $[-3, \dots, 3]$ by human annotators. Since the dataset poses the regression task, the model is optimized with $L1$ loss as task-specific objective, and we evaluate both feature-level and attention-level alignment knowledge transfer objectives $\mathcal{L}_{kt}^{cos}$ and $\mathcal{L}_{kt}^{att}$. We follow the standard protocol of the dataset and report mean average error, correlation with human annotations (annotations are obtained from multiple annotators), and 7-class accuracy. We report average results over 3 random seeds. Unimodal models are as described in Sec. 3.2 and follow the method of \cite{mult}, and the multimodal branch is identical to Figure~\ref{fig:trans}.

Table~\ref{tab:mosei} shows the results on the CMU-MOSEI dataset. Firstly, we observe that in our baseline experiments, text-only model outperforms the multimodal one (which is rather consistent with previous works, where text modality performance often lies close to the multimodal one \cite{mult}), while the text model trained under our proposed framework outperforms both of them. In fact, the proposed approach outperforms the baselines on all the modalities compared to unimodal models, with especially big increase observed in correlation metric, and the multimodal branch also outperforms the multimodal model trained independently. We observe that feature-level loss is more beneficial for improving the stronger text modality, and subsequently the multimodal branch. In turn, attention-level alignment shows to be more beneficial for audio and vision modalities. This shows that multimodal branch is mainly driven by the text modality (judging by their performance), hence features of the final hidden layer are likely to be more easily transferable to unimodal text branch than audio or vision branches. Instead, audio and vision branches can benefit from softer attention-level alignment, which does not enforce strong similarity to other modality, but instead, to tokens of the same modality enriched with multimodal information.

\textbf{Ablation studies}
We perform a few ablations on the EgoGesture dataset. First, as our primary goal is to improve the unimodal branches, we train an architecture identical to the one described earlier, but the shared weights are only updated from the uni-modal branch, and are frozen in the multimodal path. The results can be seen in Table 1, the freezing of the layers does not have a significant effect on the model, with unimodal models being marginally below the standard variant.
Next, we investigate the effect of the knowledge transfer loss and train the identical model but without optimization of the knowledge transfer objective from the multi-modal to the unimodal branch. As can be seen in Table 1, multi-modal branch still outperforms the one trained from scratch (showcasing again the benefits of unimodal gradient updates to the shared layers), but unimodal branches retain the unimodal performance, hence showing the effect of the knowledge transfer loss.
We are additionally providing ablations on the $\alpha$ (coefficient of the knowledge transfer loss), with  $\beta$ and $\gamma$ (task-specific losses) fixed to 1, which can be seen in Table 4 using EgoGesture dataset and MobileNetV2. As can be seen, any $\alpha$ outperforms the baseline, while the best result is achieved at $\alpha$=5.

\begin{table}
\footnotesize
\centering
    \begin{tabular}[b]{|cccc|}
    \hline
    \footnotesize
          & \textbf{RGB}        & \textbf{Depth}      & \textbf{MM}         \\ \hline
 {MobileNetv2 \cite{mobilenetv2}} & 86.07         & 86.67          & 87.64          \\
 \hline \hline
 {$\alpha$=1}        & 87.24         & 87.78          & 88.87          \\
 {$\alpha$=5}       & \textbf{88.57} & \textbf{88.34} & \textbf{89.19} \\
 {$\alpha$=10}      & 87.80          & 87.82          & 88.35          \\
 {$\alpha$=20}       & 87.18          & 87.36          & 88.18         \\ \hline
 \end{tabular} 
  \caption{Results with different $\alpha$ on EgoGesture}
  \vspace{-20pt}
\end{table}
\vspace{-8pt}
\section{Conclusion}

We have presented a general framework for improving performance of an arbitrary unimodal model with multimodal training that involves co-training of the unimodal models with multimodal Transformer and multi-task objective aimed at knowledge transfer from multimodal to unimodal branches. The proposed approach shows improved performance on 3 tasks of different modalities and structures. We also found that providing unimodal feedback to early layers of multimodal model aids its performance in a multimodal setting. Future work may include research on higher adaptiveness of the co-training, such that not all unimodal models are co-trained in the same manner, but instead relatively to their capacity.

\bibliographystyle{IEEEbib}
\bibliography{strings,refs}

\end{document}